# Superhuman Artificial Intelligence Can Improve Human Decision Making by Increasing Novelty

Minkyu Shin [1], Jin Kim [2], Bas van Opheusden [3], Thomas L. Griffiths [3,4]

[1] Department of Marketing, City University of Hong Kong

[2] Department of Marketing, Yale School of Management

[3] Department of Psychology, Princeton University

[4] Department of Computer Science, Princeton University

## Author Note

Minkyu Shin  https://orcid.org/0000-0001-6260-4799

Jin Kim  https://orcid.org/0000-0002-5013-3958

Bas van Opheusden  https://orcid.org/0000-0002-7169-4885

Thomas L. Griffiths  https://orcid.org/0000-0002-5138-7255

Correspondence concerning this article should be addressed to Minkyu Shin (Department of Marketing, City University of Hong Kong, Hong Kong SAR, Email: minkshin@cityu.edu.hk), Jin Kim (Department of Marketing, Yale School of Management, New Haven, CT, U.S.A., Email: jin.m.kim@yale.edu), or Bas van Opheusden (Department of Psychology, Princeton University, Princeton, NJ, U.S.A., Email: basvanopheusden@gmail.com).

**Abstract**

How will superhuman artificial intelligence (AI) affect human decision making? And what will be the mechanisms behind this effect? We address these questions in a domain where AI already exceeds human performance, analyzing more than 5.8 million move decisions made by professional Go players over the past 71 years (1950-2021). To address the first question, we use a superhuman AI program to estimate the quality of human decisions across time, generating 58 billion counterfactual game patterns and comparing the win rates of actual human decisions with those of counterfactual AI decisions. We find that humans began to make significantly better decisions following the advent of superhuman AI. We then examine human players' strategies across time and find that novel decisions (i.e., previously unobserved moves) occurred more frequently and became associated with higher decision quality after the advent of superhuman AI. Our findings suggest that the development of superhuman AI programs may have prompted human players to break away from traditional strategies and induced them to explore novel moves, which in turn may have improved their decision-making.

**Introduction**

"It made me question human creativity. When I saw AlphaGo's moves, I wondered whether the Go moves I ha[d] known were the right ones. Its style was different, and it was such an unusual experience that it took time for me to adjust. AlphaGo made me realize that I must study Go more." (1)

- Sedol Lee, a former world Go champion

Recent advances in artificial intelligence (AI) have resulted in automated systems that approach or surpass human performance in fields as diverse as medicine (e.g., diagnosing diseases) (2), transportation (e.g., autonomous driving) (3), language (e.g., ChatGPT based on GPT-3) (4), and natural sciences (e.g., AlphaFold) (5), among others (6). As AI systems outperform humans in these settings, a natural question to ask is how humans will change their own decision-making. Humans will likely adjust to such advancements in AI by delegating to, receiving aid from, or learning from AI systems to improve their own performance. But will human decision-making itself change? And what will be the mechanisms underlying this change?

Answering these questions is challenging because individual human decisions are not necessarily recorded, and AI systems are only gradually being adopted in selective domains. In this article, we overcome these challenges by studying a domain in which detailed records of human decisions are available and where the advent of superhuman AI (7) can be connected to a specific date: March 15, 2016, when AlphaGo—an AI program developed by Google's DeepMind—shocked the world by defeating a human world champion in Go. We analyze more than 5.8 million decisions made by professional Go players over the past 71 years (1950 to 2021), using a superhuman AI system to evaluate the quality of these decisions. By looking at how human play differed before and after the advent of superhuman AI[1]; we are able to evaluate its impact.

Questions about the impact of superhuman AI on human behavior are related to the literature on cumulative cultural evolution. This literature shows that there is no guarantee that human decision-making will improve in response to innovations, despite the human ability to accumulate knowledge within and across generations (8, 9). Often, cumulative cultural evolution does occur, as superior forms of decision-making are transferred from one group of individuals to another (10, 11). However, at times, intrinsic biases and frictions in human learning can delay or derail such process (12, 13). When there exist suboptimal but familiar decisions whose efficacy has been demonstrated by others, even experts fail to adopt unfamiliar but objectively better alternatives (14). It is thus not obvious whether human decision-making will improve following advancements in AI.

If innovations produced by superhuman AI do result in changes in human decision-making, a second question is what mechanism might underlie this process. Previous research suggests that novelty could be a relevant factor. It has been proposed that AI systems can generate new ideas by combining familiar ideas in novel ways and exploring conceptual spaces that may have been overlooked (15, 16). However, relatively little research has investigated whether human decision-makers will readily adopt ideas generated by an AI system; for an exception, see ref. 17.

Our analyses of professional Go players' decisions reveal that human decision-making significantly improved following the advent of superhuman AI. We also find that this improvement may be partly explained by increased novelty in human decisions made after the exogenous shock of this event. These findings illustrate that the development of superhuman AI may result in improvements in human decision-making, with innovations spreading from machines to humans and spurring further novel developments among those humans.

---

[1] We use the term "advent of superhuman AI" to denote a series of events that occurred between 2016 and 2017, including AlphaGo's victory over the human world champion in March 2016 and developments of superhuman AI programs between 2016 and 2017. These notable events are listed in SI Appendix, section 1.

## Results

We present our main results from three different sets of analyses. First, we estimate the quality of human decisions over time by using KataGo (18), a superhuman AI program. We use it to simulate billions of game patterns (i.e., 10,000 game patterns for each of the 5.8 million decisions) and compare the win rates of actual human decisions with those of counterfactual (optimal) AI decisions to construct a Decision Quality Index for each human decision (hereafter, DQI; Materials and Methods and SI Appendix, section 2.2 for details). Second, we estimate the novelty of human decisions over time by examining move sequences and identifying the first historically novel move of each game. (An additional analysis on novelty estimated from a different measure is presented in SI Appendix, Fig. S1.) Finally, we use our estimates of decision quality and novelty to estimate models testing the hypothesis that AI improved human decision making by encouraging novel decision-making.

**Decision Quality**

The top two panels of Fig. 1 present the time trends of human decision quality from 1950 to 2021. Specifically, Panels A and B show the estimated fixed effect of each year and month, respectively, on decision quality; player fixed effects were controlled for and the respective first periods were used as the baseline (i.e., 1950 for Panel A and January 1950 for Panel B). Both these panels reveal that human experts began to make significantly better decisions after the advent of superhuman AI, as evidenced by greater fixed effects of years (Panel A) and months (Panel B) after the 2016 to 2017 period than before it. Contrary to our expectations, both these panels also show that human decision-making improved comparatively little before the advent of superhuman AI (but SI Appendix, section 4.2 for minor exceptions upon closer examinations). In sum, human decision making in Go improved substantially after AlphaGo's victory in 2016 and release of superhuman AI programs in 2017, in stark contrast to the comparatively flat trend observed in the preceding 66 years (1950 to 2016).

**Figure 1**

*Estimated Fixed Effect of Each Year and Month on Decision Quality and Novelty*

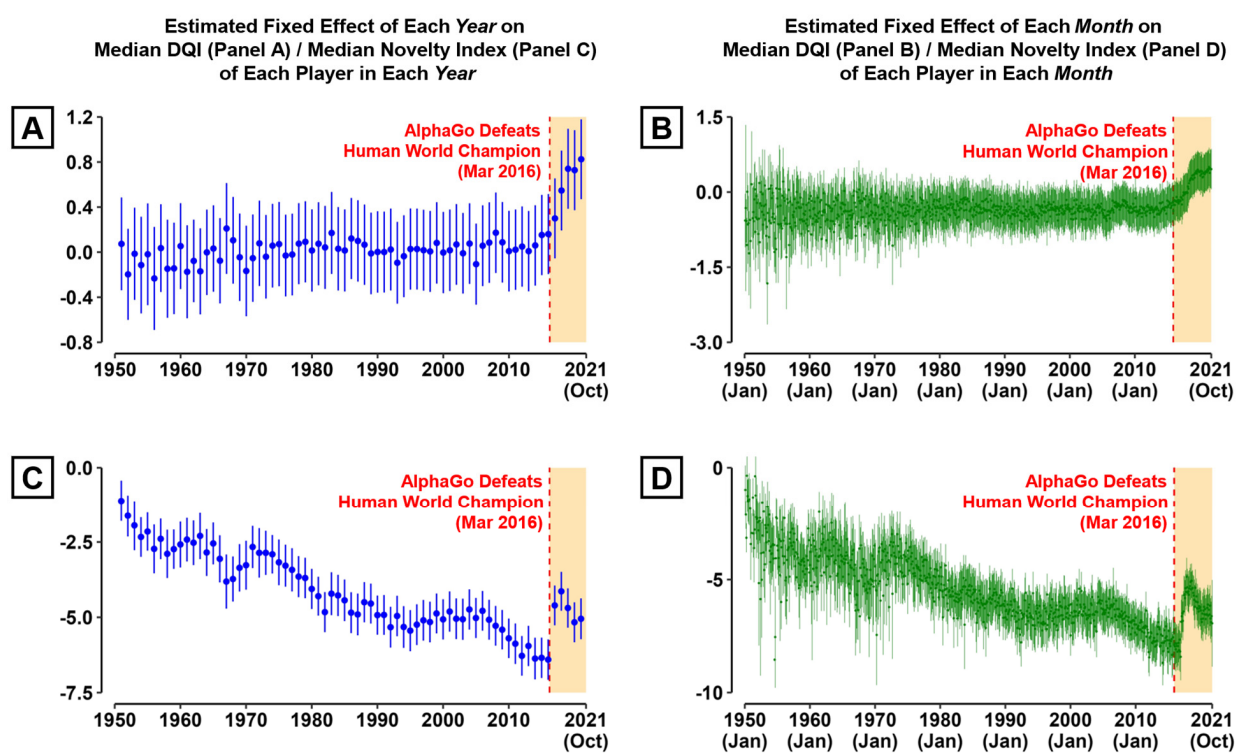

*Note.* Historical changes in quality and novelty of human decisions in Go. Panel A (Panel B) shows the fixed effect of each year (month) on decision quality along with its 95% CI, estimated using the median of Decision Quality Indices of all decisions made by each player in each year (month). Similarly, Panel C (Panel D) shows the fixed effect of each year (month) on novelty as measured with the Novelty Index, along with its 95% CI, estimated using the median of Novelty Indices of all games for each player in each year (month).

**Novelty in Decision-Making**

Next, we examined the novelty of human decisions over the same period. One way to gauge novelty in decision-making is to examine the novelty of move sequences. As is commonly done in the chess community (19), we can find in each game the first move that makes the game's move sequence historically novel (20, 21). For example, if a particular sequence of the first 9 moves in a game was observed in some game in the past, but the 10th move makes the sequence of 10 moves historically novel (i.e., never observed in any of the previous games in the dataset), then we may say that Move 10 was a novel move. If such a novel move occurs at an earlier point in one game than in another, then the earlier departure from previously observed decision-making would indicate greater novelty in the former game than in the latter. We thus used the idea of a novel move to construct a measure of novelty. Specifically, we identified the novel move in each game and subtracted its move number from the maximum move number observed in the dataset (i.e., 60[2]) to form a Novelty Index for the novel move. (SI Appendix, section 2.3 for details about the Novelty Index.) We then conducted our novelty analyses using only the data on these novel moves.

We thus estimated the novelty of human decisions over time using this Novelty Index in the same way we estimated decision quality over time using the DQI. We present the results in Fig. 1, Panels C and D. As expected, later years and months put greater downward pressure on the Novelty Index (i.e., more negative fixed effects) because the observed set of unique move sequences grew over time and pushed novel moves later. Most importantly, however, we find a sharp increase in the yearly (Panel C) and monthly (Panel D) fixed effects on the Novelty Index following the advent of superhuman AI, suggesting that players broke away from previously observed move sequences earlier in their games. An additional analysis with an alternative operationalization of novelty showed a similar increase in novelty after the advent of superhuman AI (SI Appendix, Fig. S1).

---

[2] We examined only the first 60 moves in each game because human learning from AI seems concentrated in the relatively early stage of the game (Moves 1 to 60); see ref. 22.

**Novelty and Decision Quality**

Finally, we test whether such an increase in novelty can explain the improvement in decision making. Specifically, we regressed the DQI of each move on (1) the binary variable of whether the move was made before versus after the advent of superhuman AI (After AI Dummy), (2) the binary variable of whether the move was a novel move (Novelty Dummy), and (3) their interaction. Simultaneously, we controlled for fixed effects of months (only for Model 2), move numbers, and players. Our coefficient of interest was that of the interaction term which signifies how much more novel move decisions improved as compared with non-novel move decisions, following the advent of superhuman AI. The first column in Table 1 shows the regression results not controlling for monthly fixed effects, whereas the second column shows the regression results controlling for monthly fixed effects without estimating the effect of After AI Dummy (to avoid multicollinearity).

In each of the two regression models, the interaction term was significant and positive, $\beta_3$ = 0.515. This suggests that the post-AI increase in decision quality was greater for novel moves than for non-novel moves. In other words, novel and non-novel moves together contributed to the increase in decision quality (because all moves had to be either novel or not novel, by definition), but novel moves on average contributed significantly more to the increase in decision quality than non-novel moves on average.

**Table 1**

*Decision Quality Moderated by Novelty*

| Dependent Variable: Decision Quality Index (DQI) | *Model 1* | *Model 2* |
|---|---|---|
| After AI Dummy ($\beta_1$) | 0.59754*** (0.01601) | – |
| Novelty Dummy ($\beta_2$) | – 0.60770*** (0.01219) | – 0.60777*** (0.01218) |
| After AI Dummy × Novelty Dummy ($\beta_3$) | 0.51504*** (0.02147) | 0.51498*** (0.02132) |
| Monthly fixed effects | No | Yes |
| Move number fixed effects | Yes | Yes |
| Player fixed effects | Yes | Yes |
| Observations | 5,857,513 | 5,857,513 |

*Note.* Standard errors are clustered at player level; *** $p < .001$

**Testing an Alternative Explanation: Memorization**

One possible hypothesis that may explain the post-AI increase in both decision quality and novelty is that human players simply memorized and replicated the superior and novel decisions produced by the AI systems without any internalization of the AI's decision-making logic (hereafter, the memorization hypothesis). Such memorization can take the form of either memorizing sequences of AI decisions (especially from the beginning of the game) or memorizing an optimal AI decision for each likely state. In this section, we present evidence that memorization is unlikely to fully explain the post-AI increase in decision quality and novelty. Specifically, in the first three subsections, we show that neither of the aforementioned forms of memorization could fully account for the increase in decision quality. In the last two subsections, we show that memorization of sequences of AI decisions is unlikely to explain the increase in novelty.

**The Quality of Human Decisions That Differ from the Optimal AI Decisions**

One way to test the memorization hypothesis is to examine the move decisions that differ from the counterfactual (optimal) AI decisions. If memorization of the optimal AI decisions fully explains the increase in decision quality, then human decisions that differed from the optimal AI decisions would not show the same increase in quality after the advent of superhuman AI. On the other hand, if memorization cannot fully explain the post-AI increase in decision quality, then human decisions that differed from the optimal AI decisions could still show the post-AI increase in quality.

With this reasoning, we estimated the time trends of decision quality only for the decisions that differed from the optimal AI decisions. In our dataset, about 40% of all human decisions matched the optimal AI decisions, leaving us with about 60% of all human decisions for this analysis.

Fig. 2 shows the time trends of decision quality estimated from only the move decisions that differed from counterfactual (optimal) AI move decisions. As with the previous analysis (Fig. 1, Panel A), we still find a similar increase in the yearly fixed effects on decision quality after the advent of superhuman AI, even when our analysis focused on the move decisions that differed from those of AI. Estimating the monthly fixed effects for this restricted set of decisions shows the same post-AI increase in decision quality (SI Appendix, Fig. S2).

As this analysis involved only the decisions that differed from the optimal AI decisions, memorization of the optimal AI decisions cannot explain the still-evident increase in decision quality following the advent of superhuman AI.

**Figure 2**

*Yearly Fixed Effect on Decision Quality for Decisions That Differ From AI Decisions*

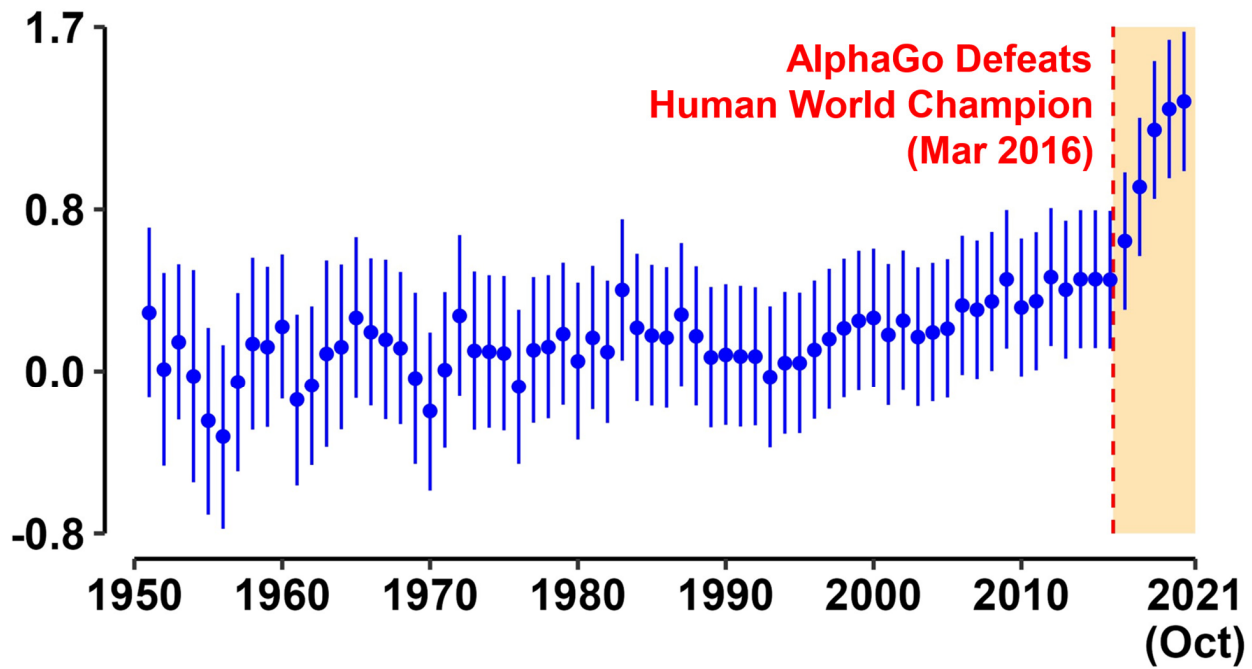

*Note.* Yearly fixed effects on DQI, constructed from a dataset that excludes the move decisions that matched the counterfactual (optimal) decisions of the AI program. We find a similar increase in human decision quality following the advent of superhuman AI. The upward time trends of DQI after 2016 are still evident even when analyzing the restricted set of move decisions. This suggests that memorization hypothesis cannot fully explain the post-AI increase in decision quality.

**The Quality of Human Decisions in Later Stages of the Game**

As another way to test the memorization hypothesis, we analyzed the quality of decisions at different stages of the game, focusing especially on relatively later stages of the game (e.g., Moves 21 to 60). Because memorization of the optimal AI decisions in these later stages would be practically impossible (due to the complexity of Go), a post-AI increase in decision quality also in these stages would be evidence against the memorization hypothesis.

Prior to conducting these analyses, we found that more than half of all novel moves in our entire dataset occur before Move 10 and that 99% of opening move sequences become historically novel by Move 21. This meant that moves after Move 20 would likely present a completely novel state of the board to human players (i.e., different from every state they have ever faced), and that memorizing the optimal move for the completely novel state would not have been feasible. Based on this knowledge of novel moves, we split our dataset into six subsets of 10 moves and examined the time trends of decision quality for each set of 10 moves.

If the human decision quality increased because the humans simply memorized series of optimal decisions generated by an AI program, then the post-AI increase in decision quality would likely be observed for the earliest stages of the game (e.g., Moves 1 to 10 and Moves 11 to 20) but would unlikely be observed for relatively later stages of the game (e.g., Moves 21 to 60). On the other hand, if the human decision quality increased because the humans were making better decisions themselves, then the post-AI increase in decision quality would be observed in both types of stages (i.e., not only for Moves 1 to 20 but also for Moves 21 to 60). Indeed, our data are more consistent with this latter case (SI Appendix, Fig. S3). That is, human decision quality increased not only for the earliest stages of the game (Moves 1 to 10‡ and Moves 11 to 20) but also for the relatively later stages of the game (Moves 21 to 50); SI Appendix, section 4.2 for more discussion on this analysis. These results provide additional evidence against the memorization hypothesis.

**The Quality of Decisions After the Opponent Deviates from the Optimal AI Decisions**

We further test the memorization hypothesis by examining the move decisions made after opponents deviate from a series of optimal AI decisions. Suppose players and their opponents memorized sequences of optimal AI decisions, and such memorization fully explains the post-AI increase in decision quality. If so, in cases where an opponent first deviates from the series of optimal AI decisions, the focal player's decision in response would not be part of the memorized sequence and would be unlikely to show the post-AI increase in quality. Thus, examining such decisions allows another test of the memorization hypothesis.

To illustrate, consider the following move sequence: *dd* (Black) → *pp* (White) → *dp* (Black) → *pd* (White). If the first two decisions (*dd* and *pp*) matched the optimal AI decisions, but the next decision by the opponent, *dp*, did not match the optimal AI decision, then the decision made in response to the opponent's deviation, namely *pd*, would be analyzed.

We thus repeated the analysis of decision quality as in the Results section but focusing only on move decisions that come immediately after the opponent deviates from the optimal AI decision. Indeed, we do find that decisions in response to the opponent's deviation also show an increase in quality following the advent of superhuman AI (SI Appendix, Fig. S4) and SI Appendix, section 4.3 for more discussion on this analysis. This is inconsistent with the memorization hypothesis, suggesting that the increase in human decision quality following the advent of superhuman AI is unlikely to be a result of human players simply memorizing sequences of AI decisions.

**Re-estimating Time Trends of Novelty Index for Novel Move Decisions that Do or Do Not Match the Optimal AI Decisions**

In this and the following subsections, we test whether the memorization hypothesis can fully account for the post-AI increase in novelty of human decisions. Recall that after the advent of superhuman AI, the yearly and monthly fixed effects on Novelty Index increased (Fig. 1 C and D). In other words, the historically novel move in each game tended to occur earlier in the game after the advent of superhuman AI. So, if novel moves occurring earlier in the game is simply a result of humans memorizing and replicating the optimal AI decisions, then examining only the novel moves that differed from the optimal AI decisions would not show such evidence for the increase in novelty.

With the reasoning above, we estimated the yearly fixed effects on the Novelty Index analyzing only the novel move decisions that differed from the corresponding optimal AI decisions (Fig. 3). A comparison with the temporal pattern from the move decisions that match the optimal AI decisions is also provided (SI Appendix, Fig. S5). In both analyses, we find the upward shift in the time trends of the Novelty Index following the advent of superhuman AI. Even when we examine only the novel move decisions that differed from the optimal AI decisions (i.e., decisions in which the corresponding optimal AI decisions were not memorized), the upward shift in Novelty Index is readily observable. This suggests that the increase in novelty cannot be explained solely by humans memorizing and replicating the optimal AI decisions.

**Figure 3**

*Yearly Fixed Effect on Novelty for Decisions That Differ From AI Decisions*

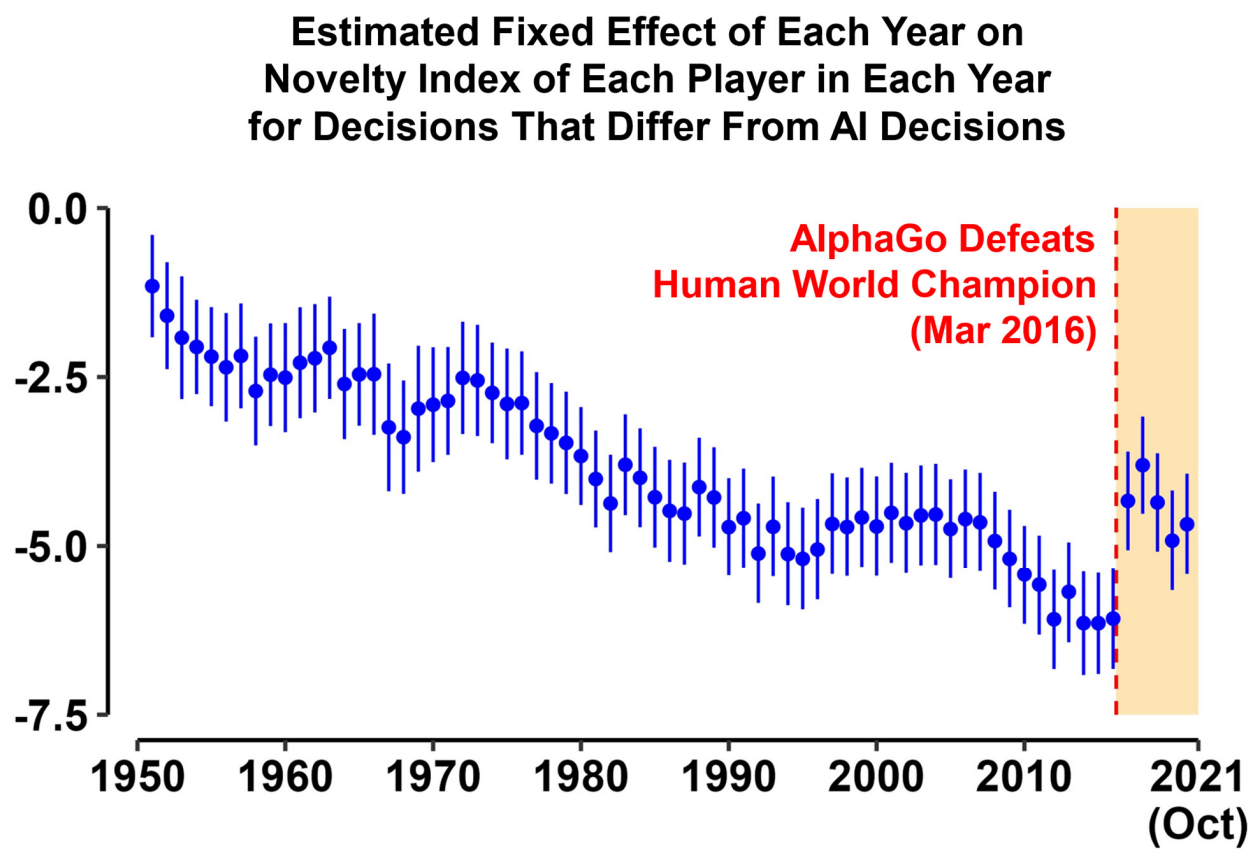

*Note.* Yearly fixed effects on Novelty Index, constructed from a dataset that excludes the move decisions that matched the counterfactual (optimal) decisions of the AI program. Even when this restricted set of move decisions were analyzed, we still find a similar increase in novelty following the advent of superhuman AI.

**Re-estimating Time Trends of Novelty Index After Adding AI Decisions into the Dataset**

As yet another test of the memorization hypothesis regarding the post-AI increase in decision novelty, we repeated the analyses of novelty in Results section (Fig. 1 C and D) after adding 600k move decisions made by a superhuman AI program into the dataset. For the previous analyses of the Novelty Index, we had identified the historically novel move in each game by considering only the previously observed sequences of human decisions. As a result, if humans witnessed historically novel decisions in their private studies with AI programs and then they themselves made the same decisions, such decisions would be considered historically novel (human) decisions in our analyses, even though they were initially discovered by AI programs. If such memorization occurred often enough, a substantial proportion of novel moves may have been misattributed to novelty by humans rather than novelty by AI programs.

To test for this possibility, we generated sequences of AI move decisions (i.e., AI versus AI) and inserted them in our dataset, temporally positioning them immediately before the advent of superhuman AI. We then redetermined novel moves with this expanded set of decision sequences artificially inserted and positioned prior to the advent of superhuman AI. If humans were merely memorizing novel decisions initially discovered by the AI programs, such decisions would now be less likely to be counted as historically novel because the same decisions would likely already exist in the dataset as having been made before the advent of superhuman AI. Thus, if we find the same post-AI increase in novelty (as measured with the Novelty Index) even after adding the simulated AI decisions, it would present further evidence against the hypothesis that memorization of AI decisions increased novelty.

Indeed, that is exactly what we find (SI Appendix, Fig. S6). Even when the 600k newly simulated AI decisions were added to the dataset as having been made before the advent of superhuman AI, and the Novelty Index of each game was recalculated accordingly, the time trends of the Novelty Index hardly change. In other words, the increase in novelty is still evident even when accounting for the possibility that human players may have used AI programs to generate and memorize promising game patterns.

## Discussion

As AI systems continue to approach or surpass human abilities in various fields, it is essential to comprehend the effects they have on human decision-making (23–26). This research topic, specifically within the context of the game of Go, primarily has centered on assessing the change in human decision quality. The increase in human decision quality following the advent of superhuman AI was initially documented by the first two authors of this article (22) and was subsequently corroborated by other research groups (27). However, to our knowledge, none of this research simultaneously investigated changes in decision novelty and linked them with the increase in decision quality.

In this research, we find that human decision-making significantly improved following the advent of superhuman AI and that this improvement was associated with greater novelty in human decisions. Because AI can identify optimal decisions free of human biases (especially when it is trained via self-play), it can ultimately unearth superior solutions previously neglected by human decision-makers who may be focused on familiar solutions. The discovery of such superior solutions creates opportunities for humans to learn and innovate further.

One important question concerns how much of the observed increase in decision quality and novelty can be attributed to players internalizing the AI systems' superior decision-making logic as opposed to merely memorizing AI decisions. Our analyses both in the main text and SI Appendix show that memorization of AI decisions cannot be the sole explanation for the increase in decision quality and novelty. For example, when we exclude from our analysis all human decisions that matched the optimal AI decisions and examine the quality of the remaining human decisions, we still find a sharp increase in decision quality after the advent of superhuman AI.

Our findings raise interesting questions for future research, including 1) whether the advent of superhuman AI increased novelty and thereby increased decision quality (i.e., whether each link in the possible causal chain can be established), 2) through which mechanism(s) superhuman AI increased novelty and decision quality (if not through novelty), 3) how the historically novel decisions themselves qualitatively differed before versus after the advent of superhuman AI, and 4) how styles of decision-making changed after the advent of superhuman

AI. There are numerous other related questions worth examining, and we hope that our findings can encourage such investigations not only in contexts similar to Go but also in other contexts that allow careful examination of human decision-making. The increasing availability of superhuman AI systems opens exciting new frontiers for studying human cognition (28).

## Materials and Methods

### Data

Raw data came from Games of Go on Disk ("GoGoD"), a copyrighted collection of Go data (29) on move-by-move decisions. We purchased the database (i.e., zipped sgf files) and created a dataset for our analyses (29-32). We then used this dataset to construct the Decision Quality Index (DQI) and Novelty Index. Calculating the DQI for each move decision required a superhuman AI program (18) to generate the AI-optimal move decision and estimate the difference between the win rate of the counterfactual (optimal) AI decision and that of the actual human decision. Calculating the Novelty Index for each game did not require any superhuman AI program. More details about both measures are presented in SI Appendix, section 2.

### Estimating Decision Quality Index (DQI)

To use observational data to track the quality of human decisions over time, it is essential to establish an objective measure that quantifies the quality of human decisions. However, the large state space of Go makes it difficult to determine which decisions are objectively better than others. We thus leverage a superhuman AI program KataGo (18) that peers into the complex relationship between a decision made in a given state and the final outcome of the game (i.e., win or lose) associated with the decision.

To construct the Decision Quality Index (DQI), we take advantage of two features of the AI program: 1) It makes decisions of superhuman quality, and 2) it can evaluate the quality of any decision, whether the decision was made by a human player or itself. We first enter into the AI program the game log data for all games between professional Go players. The log data contain a series of move decisions made by human players. Then, for each move decision, we have the AI program make its own optimal decision in the same state. This AI-generated decision serves as a counterfactual (optimal) decision (Step 1). Using the same AI program, we compare the win rate associated with the AI-generated decision with the win rate associated with the human decision (Step 2). The DQI of a given human decision, then, equals [100% - (win rate of the Counterfactual AI Decision - win rate of the actual human decision)]. SI Appendix, section 2.2 has a more technical definition.

**Data, Materials, and Software Availability**

Replication data with AI simulation output to calculate the DQI are available on the project's page in Open Science Framework (OSF) at https://osf.io/xpf3q/. The codes for constructing the DQI and replicating figures and tables are also available on the OSF page. Datasets have been deposited at https://osf.io/xpf3q/files/osfstorage.

THE IMPACT OF AI ON DECISION QUALITY AND NOVELTY IN GO                              2223. Y. Alufaisan, L. R. Marusich, J. Z. Bakdash, Y. Zhou, M. Kantarcioglu, Does explainable artificial intelligence improve human decision-making? Proc. AAAI Conf. Artif. Intell. 35, 6618–6626 (2021).

24. T. M. Rawson, R. Ahmad, C. Toumazou, P. Georgiou, A. H. Holmes, Artificial intelligence can improve decision-making in infection management. Nat. Hum. Behav. 3, 543–545 (2019).

25. F. Callaway et al., Leveraging artificial intelligence to improve people's planning strategies. Proc. Natl. Acad. Sci. U.S.A. 119, e2117432119 (2022).

26. F. Becker, J. Skirzyn´ski, B. van Opheusden, F. Lieder, Boosting human decision-making with AI-generated decision aids. arXiv [Preprint] (2022). https://arxiv.org/abs/2203.02776

27. S. Choi, N. Kim, J. Kim, H. Kang, How Does AI Improve Human Decision-Making? Evidence from the AI-Powered Go Program, USC Marshall School of Business Research Paper (2022).

28. D. Purves, What does AI's success playing complex board games tell brain scientists? Proc. Natl. Acad. Sci. U.S.A. 116, 14785–14787 (2019).

29. T. M. Hall, J. Fairbairn, Games of Go on Disk Database (2021). https://gogodonline.co.uk

30. M. Shin, J. Kim, B. Van Opheusden, T. L. Griffiths, Supporting Information for Superhuman artificial intelligence can improve human decision-making by increasing novelty (2023). https://doi.org/10.17605/OSF.IO/5CDMA

31. J. Kim, 'kim': A Toolkit for Behavioral Scientists (2022). https://jinkim.science/docs/kim.pdf

32. B. A. Beheim, kaya Go game analysis (2022). https://github.com/babeheim/kaya

33. B. A. Beheim, C. Thigpen, R. McElreath, Strategic social learning and the population dynamics of human behavior: The game of Go. Evol. Hum. Behav. 35, 351–357 (2014).
The published version of this article is available at PNAS: https://doi.org/10.1073/pnas.2214840120